\documentclass[11pt]{article}
\usepackage[numbers]{natbib}
\usepackage{fullpage}
\usepackage{hyperref}
\usepackage{url}

\usepackage{pgfplotstable}
\usepackage{graphicx}
\usepackage{float}

\providecommand{\comment}[1]{}

\usepackage{statistics-macros}
\usepackage{subfigure}
\usepackage{color}
\usepackage{algorithm}
\usepackage{algpseudocode}
\usepackage{tabularx}

\usepackage{manyfoot}

\DeclareNewFootnote{A}
\DeclareNewFootnote{B}

\let\footnoteR\footnoteB
\let\footnote\footnoteA

\usepackage{caption}
\usepackage{graphicx}

\definecolor{matlab_blue}{rgb}{0,    0.4470,   0.7410}
\definecolor{matlab_red}{rgb}{0.8500,    0.3250,    0.0980}
\definecolor{matlab_yellow}{rgb}{0.9290,    0.6940,    0.1250}
\definecolor{matlab_purple}{rgb}{0.4940,    0.1840,    0.5560}
\definecolor{matlab_green}{rgb}{0.4660,    0.6740,    0.1880}

\title{Scalable End-to-End Autonomous Vehicle \\
  Testing via Rare-event Simulation}

\makeatletter
\long\def\@makecaption#1#2{
  \vskip 0.8ex
  \setbox\@tempboxa\hbox{\small {\bf #1:} #2}
  \parindent 1.5em  %
  \dimen0=\hsize
  \advance\dimen0 by -3em
  \ifdim \wd\@tempboxa >\dimen0
  \hbox to \hsize{
    \parindent 0em
    \hfil 
    \parbox{\dimen0}{\def\baselinestretch{0.96}\small
      {\bf #1.} #2
    } 
    \hfil}
  \else \hbox to \hsize{\hfil \box\@tempboxa \hfil}
  \fi
}
\makeatother

\begin{document}
\abovedisplayskip=8pt plus0pt minus3pt
\belowdisplayskip=8pt plus0pt minus3pt

\begin{center}
  {\LARGE Scalable End-to-End Autonomous Vehicle Testing\\ via Rare-event Simulation} \\
  \vspace{.5cm} {\Large Matthew O'Kelly\footnoteR{Equal contribution}$^{1}$~~Aman Sinha$^{*2}$~~Hongseok Namkoong$^{*2}$\\ John Duchi$^{2}$~~Russ Tedrake$^{3}$}\\
  \vspace{.2cm}
  $^{1}${\large University of Pennsylvania}\\
  $^2${\large Stanford University}\\
  $^3${\large Massachusetts Institute of Technology}\\
  \vspace{.2cm} \texttt{mokelly@seas.upenn.edu \{amans,hnamk,jduchi\}@stanford.edu russt@mit.edu}
\end{center}

\newif\ifarxiv
\arxivtrue
\begin{abstract}

While recent developments in autonomous vehicle (AV) technology highlight
substantial progress, we lack tools for rigorous and scalable testing. Real-world
testing, the \textit{de facto} evaluation environment, places the public in
danger, and, due to the rare nature of accidents, will require billions of
miles in order to statistically validate performance claims. We implement a simulation
framework that can test an entire modern autonomous driving system, including, in particular, systems that employ deep-learning perception and control algorithms. %
Using adaptive importance-sampling methods to accelerate rare-event probability evaluation, we estimate the probability of an accident under a base distribution governing standard traffic behavior. We demonstrate our framework on a highway scenario, accelerating system
evaluation by $2$-$20$ times over naive
Monte Carlo sampling methods and $10$-$300 \mathsf{P}$ times (where $\mathsf{P}$ is the number of processors) over real-world testing.

 \end{abstract}

\section{Introduction}
\label{sec:intro}

Recent breakthroughs in deep learning have accelerated the development of
autonomous vehicles (AVs); many research prototypes now operate %
on real roads alongside human drivers.  While advances in computer-vision
techniques have made human-level performance possible on narrow perception
tasks such as object recognition, several fatal accidents involving AVs
underscore the importance of testing whether the perception and control
pipeline---when considered as a \emph{whole system}---can safely interact with
humans. Unfortunately, testing AVs in real environments, the most
straightforward validation framework for system-level input-output behavior,
requires prohibitive amounts of time due to the rare nature of serious
accidents \cite{shalev2017formal}. Concretely, a recent study~\cite{KalraPa16}
argues that AVs need to drive ``hundreds of millions of miles and, under some
scenarios, hundreds of billions of miles to create enough data to clearly
demonstrate their safety.''
Alteratively, formally verifying an AV algorithm's
``correctness''~\cite{KwiatkowskaNoPa11,Althoff2014, SeshiaSaSa15, apex_SAE16}
is difficult since all driving policies are subject to crashes
caused by other drivers~\cite{shalev2017formal}. It is unreasonable to ask
that the policy be safe under \emph{all} scenarios.
Unfortunately, ruling out scenarios where the AV should not be blamed is a task
subject to logical inconsistency, combinatorial growth in specification
complexity, and subjective assignment of fault.

Motivated by the challenges underlying real-world testing and formal
verification, we consider a probabilistic paradigm---which we call a
\emph{risk-based framework}---where the goal is to evaluate the
\emph{probability of an accident} under a base distribution representing
standard traffic behavior. By assigning learned probability values to
environmental states and agent behaviors, our risk-based framework considers
performance of the AV's policy under a data-driven model of the world. To
efficiently evaluate the probability of an accident, we implement
a photo-realistic and physics-based simulator that provides the AV with
perceptual inputs (\eg~video and range data) and traffic conditions (\eg~other
cars and pedestrians). The simulator allows parallelized, faster-than-real-time evaluations in varying environments (\eg~weather, geographic locations, and aggressiveness of other cars).

Formally, we let $P_0$ denote the base distribution that models standard
traffic behavior and $X \sim P_0$ be a realization of the simulation
(\eg~weather conditions and driving policies of other agents). For an
objective function $f: \mc{X} \to \R$ that measures ``safety''---so that low
values of $f(x)$ correspond to dangerous scenarios---our goal is to evaluate
the probability of a dangerous event
\begin{equation}
  \label{eqn:rare-prob}
  p_{\gamma} \defeq \P_0(\obj(X) \le \gamma)
\end{equation}
for some threshold $\gamma$. Our risk-based framework is agnostic to the
complexity of the ego-policy and views it as a black-box module. Such
an approach allows, in particular, deep-learning based perception systems that
make formal verification methods intractable.

An essential component of this approach is to estimate the base distribution
$P_0$ from data; we use public traffic data collected by the US Department of
Transportation~\cite{Ngsim08}. While such datasets do not offer insights into
how AVs interact with human agents---this is precisely why we design
our simulator---they illustrate the range of standard human driving behavior
that the base distribution $P_0$ must model. We use imitation
learning~\cite{Russell98,RossBa10,RossGoBa11,HoEr16,BaramAnOrCasMa17} to learn
a generative model for the behavior (policy) of environment vehicles; unlike
traditional imitation learning, we train an ensemble of models to characterize
a distribution of human-like driving policies.

As serious accidents are rare ($p_{\gamma}$ is small), we view this as a
\emph{rare-event simulation}~\cite{AsmussenGl07} problem; naive Monte Carlo
sampling methods require prohibitively many simulation rollouts to generate
dangerous scenarios and estimate $p_{\gamma}$. To accelerate safety
evaluation, we use adaptive importance-sampling methods to learn alternative
distributions $P_{\theta}$ that generate accidents more
frequently. Specifically, we use the cross-entropy
algorithm~\cite{RubinsteinKr04} to iteratively approximate the optimal
importance sampling distribution. In contrast to simple classical
settings~\cite{RubinsteinKr04, ZhaoHuPeLaLe18} which allow analytic updates to
$P_{\theta}$, our high-dimensional search space requires solving convex
optimization problems in each iteration
(Section~\ref{sec:risk}). %
To address numerical instabilities of importance sampling estimators in high dimensions,
we carefully design search spaces and perform computations in logarithmic scale. Our implementation produces $2$-$20$ times as many rare
events as naive Monte Carlo methods, independent of the complexity of the
ego-policy.

In addition to accelerating evaluation of $p_{\gamma}$, learning a
distribution $P_{\theta}$ that \emph{frequently} generates realistic dangerous
scenarios $X_i \sim P_{\theta}$ is useful for engineering purposes. The
importance-sampling distribution $P_{\theta}$ not only efficiently samples
dangerous scenarios, but also ranks them according to their likelihoods under
the base distribution $P_0$. This capability
enables a deeper understanding of failure modes and prioritizes their
importance to improving the ego-policy.

\begin{figure}[t]
  \centering
	\includegraphics[width=.7\textwidth]{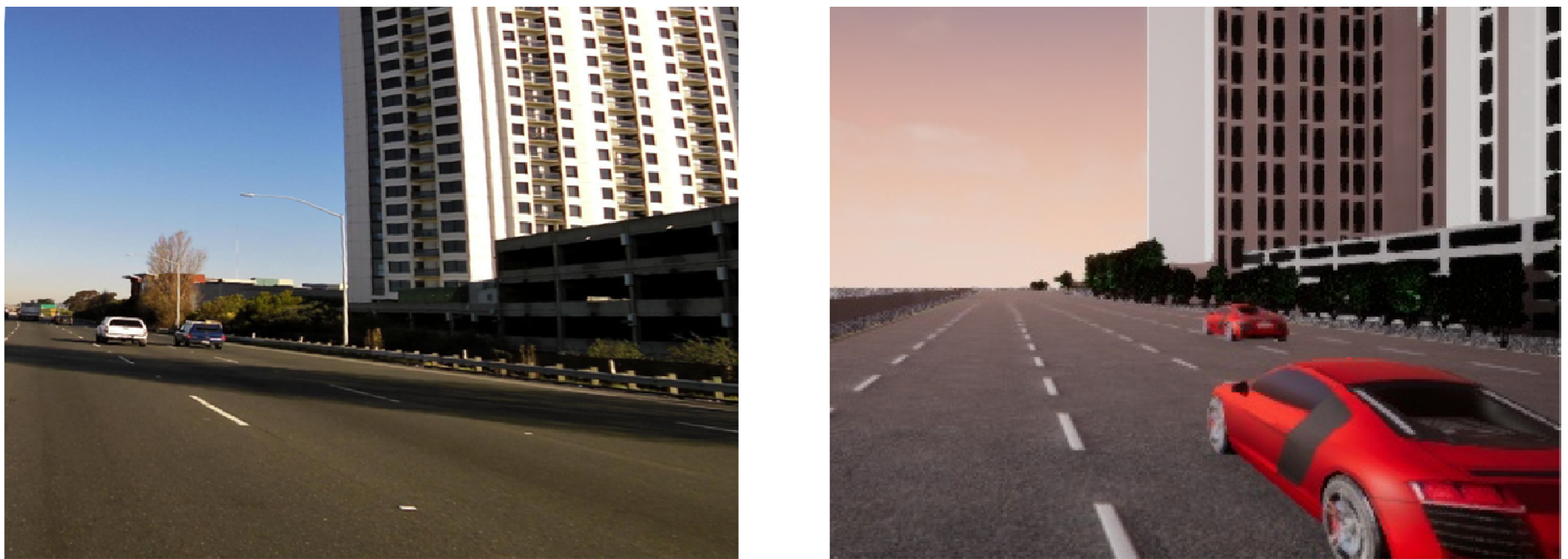}
	\caption{Multi-lane highway driving on I-80: (left) real image, (right) rendered
          image from simulator}
	\label{fig:i80scenario}
\end{figure}

As a system, our simulator allows fully distributed rollouts, making our
approach orders of magnitude cheaper, faster, and safer than real-world
testing. Using the asynchronous messaging library ZeroMQ~\cite{Hintjens13},
our implementation is fully-distributed among
available CPUs and GPUs; our rollouts are up to $30 \mathsf{P}$
times faster than real time, where $\mathsf{P}$ is the number of
processors. Combined with the cross-entropy method's speedup, we achieve $10$-$300 \mathsf{P}$ speedup over real-world testing. 

In what follows, we describe components of our open-source toolchain, a
photo-realistic simulator equipped with our data-driven risk-based framework
and cross-entropy search techniques. The toolchain can test an AV as a
\emph{whole system}, simulating the driving policy of the ego-vehicle by
viewing it as a black-box model. The use of adaptive-importance sampling
methods motivates a unique simulator architecture
(Section~\ref{sec:simulator}) which allows real-time updates of the policies
of environment vehicles. In Section~\ref{sec:experiments}, we test our
toolchain by considering an end-to-end deep-learning-based
ego-policy~\cite{bojarski2016end} in a multi-agent highway
scenario. Figure~\ref{fig:i80scenario} shows one configuration of this
scenario in the real world along with rendered images from the simulator,
which uses Unreal Engine 4 \cite{games2015unreal}. Our experiments show that we accelerate the assessment of rare-event probabilities with respect to naive Monte Carlo methods as well as real-world testing. 
We believe
our open-source framework is a step towards a rigorous yet scalable
platform for evaluating AV systems, with the broader goal of understanding how
to reliably deploy deep-learning systems in safety-critical applications.

\section{Rare-event simulation}
\label{sec:risk}

To motivate our risk-based  framework, we first argue that
formally verifying correctness of a AV system is infeasible due to the challenge of defining ``correctness.'' Consider a scenario where an AV commits a traffic violation to avoid collision with an out-of-control truck approaching from behind. If the ego-vehicle decides to avoid collision by running through a red light with no further ramifications, is it ``correct'' to do so? The ``correctness'' of the policy depends on
the extent to which the traffic violation endangers nearby humans and
whether any element of the ``correctness'' specification explicitly forbids
such actions. That is, ``correctness'' as a binary output is a concept defined by its exceptions, many elements of which are subject to individual valuations~\cite{Bonnefon1573}.

Instead of trying to verify correctness, we begin with a continuous measure
of safety $\obj: \mc{X} \to \R$, where $\mc{X}$ is space of traffic conditions
and behaviors of other vehicles. The prototypical example
in this paper is the minimum
time-to-collision (TTC) (see Appendix \ref{section:scenario} for its definition) to other environmental agents over a simulation
rollout. Rather than requiring safety for all $x \in \mc{X}$, we relax the
deterministic verification problem into a probabilistic one
where we are concerned with the probability under standard traffic conditions that $\obj(X)$ goes below a safety threshold. Given a
distribution $P_0$ on $\mc{X}$, our goal is to estimate the rare
event probability $p_{\gamma} \defeq P_0(\obj(X) \le \thresh)$ based on
simulated rollouts $\obj(X_1), \ldots, \obj(X_n)$. As accidents are rare
and $p_{\gamma}$ is near $0$, we treat this as a rare-event
simulation problem; see~\cite[Chapter VI]{Bucklew13,AsmussenGl07}
for an overview of this topic.

First, we briefly illustrate the well-known difficulty
of naive Monte Carlo simulation when
$p_{\gamma}$ is small. From a sample $X_i \simiid P_0$, the naive
Monte Carlo estimate is
$\what{p}_{N, \thresh} \defeq \frac{1}{N} \sum_{i=1}^N \indic{\obj(X_i) \le
  \thresh}$. As $p_{\thresh}$ is small, we use relative accuracy
to measure our performance, and the central limit theorem implies
the relative accuracy is approximately
\begin{equation*}
  \left| \frac{\what{p}_{N, \thresh}}{p_{\thresh}} - 1\right|
  \stackrel{\rm dist}{\approx}
  \sqrt{\frac{1-p_{\thresh}}{N p_{\thresh}}} \left|Z\right|
  + o(1/\sqrt{N})
  ~~ \mbox{for}~ Z \sim \normal(0, 1).
\end{equation*}
For small $p_\thresh$, we require a sample of size
$N \gtrsim 1 / (p_\thresh \epsilon^2)$
to achieve $\epsilon$-relative accuracy, and if $\obj(X)$ is
light-tailed, the sample size must grow exponentially
in $\thresh$.

\paragraph{Cross-entropy method}

As an alternative to a naive Monte Carlo estimator, we consider (adaptive)
importance sampling~\cite{AsmussenGl07}, and we use a model-based optimization
procedure to find a good importance-sampling distribution. The optimal
importance-sampling distribution for estimating $p_{\thresh}$ has the
conditional density
$p\opt(x) = \indic{\obj(x) \le \thresh} p_0(x) / p_{\thresh}$, where $p_0$ is
the density function of $P_0$: as $p_0(x) / p\opt(x) = p_{\thresh}$ for all
$x$ satisfying $\indic{\obj(x) \le \thresh}$, the estimate
$\what{p}_{N, \thresh}\opt \defeq \frac{1}{N} \sum_{i=1}^N
\frac{p_0(X_i)}{p\opt(X_i)} \indic{\obj(X_i) \le \thresh}$ is
exact. This sampling scheme is, unfortunately,
\emph{de facto} impossible, because
we do not know $p_{\gamma}$. Instead, we use a parameterized importance
sampler $P_{\theta}$ and employ an iterative model-based search method to
modify $\theta$ so that $P_\theta$ approximates $P\opt$.

The cross-entropy method~\cite{RubinsteinKr04} iteratively tries to find
$\theta\opt \in \argmin_{\theta \in \Theta} \dkl{P\opt}{P_{\theta}}$, the
Kullback-Leibler projection of $P\opt$ onto the class of parameterized
distributions $\mc{P} = \{ P_{\theta} \}_{\theta \in \Theta}$.  Over
iterations $k$, we maintain a surrogate distribution
$q_k(x) \propto \indic{\obj(x) \le \gamma_k} p_0(x)$ where
$\gamma_k \ge \gamma$ is a (potentially random) proxy for the rare-event
threshold $\gamma$, and we use samples from $P_{\theta}$ to update $\theta$ as
an approximate projection of $Q$ onto $\mc{P}$. The motivation underlying this
approach is to update $\theta$ so that $P_{\theta}$ upweights regions of
$\mc{X}$ with low objective value (\ie~unsafe) $\obj(x)$. We fix a quantile
level $\tol \in (0, 1)$---usually we choose $\tol \in [0.01, 0.2]$---and use the
$\tol$-quantile of $f(X)$ where $X \sim P_{\theta_k}$ as $\thresh_k$, our
proxy for the rare event threshold $\thresh$ (see~\cite{HomemDeMello07} for
alternatives). We have the additional challenge that the $\rho$-quantile of
$\obj(X)$ is unknown, so we approximate it using i.i.d.\ samples
$X_i \sim P_{\theta_k}$.  Compared to applications of the
cross-entropy method~\cite{RubinsteinKr04, ZhaoHuPeLaLe18} that focus on
low-dimensional problems permitting analytic updates to $\theta$, our
high-dimensional search space requires solving convex optimization problems in
each iteration. To address numerical challenges in computing likelihood
ratios in high-dimensions, our implementation carefully constrains the search space 
and we compute likelihoods in logarithmic scale.

We now rigorously describe the algorithmic details. First, we use natural exponential families as our class of 
importance samplers $\mc P$.

\begin{algorithm}[t!]
  \caption{Cross-Entropy Method}
  \label{alg:ce}
  \begin{algorithmic}[1]
    \State Input: Quantile $\tol \in (0, 1)$,
    Stepsizes $\{\alpha_k\}_{k \in \N}$, Sample sizes $\{N_k \}_{k \in \N}$,
    Number of iterations $K$
    \State Initialize: $\theta_0 \in \Theta$
    \State \textbf{for} $k = 0, 1, 2, \dots, K-1$~\textbf{do}
    \State~~~~ Sample $X_{k, 1}, \ldots, X_{k, N_k} \simiid P_{\theta_k}$
    \State~~~~ Set $\gamma_k$ as the minimum of $\gamma$ and the
    $\tol$-quantile of
    $f(X_{k, 1}), \ldots, f(X_{k, N_k})$
    \State~~~~
      $\theta_{k+1}
      = \argmax_{\theta \in \Theta} \left\{
    \alpha_k \theta^\top D_{k+1}
    + (1-\alpha_k) \theta^\top \nabla A(\theta_k) - A(\theta) \right\}$
  \end{algorithmic}
\end{algorithm}

\begin{definition}
  The family of density functions $\{p_{\theta}\}_{\theta \in \Theta}$,
  defined with respect to base measure $\mu$,
  is a \emph{natural exponential family} if there exists a sufficient
  statistic $\suff$ such that $p_{\theta}(x) =
  \exp(\theta^\top \suff(x) - A(\theta))$ where $A(\theta) = \log
  \int_{\mc{X}} \exp(\theta^\top \suff(x)) d\mu(x)$ is the log partition
  function and $\Theta \defeq \{ \theta \mid A(\theta) < \infty\}$.
\end{definition}
\noindent Given this family, we consider idealized updates to the parameter
vector $\theta_k$ at iteration $k$, where we compute projections of a mixture
of $Q_k$ and $P_{\theta_k}$ onto $\mc{P}$
\begin{align}
  \theta_{k+1}
  & = \argmin_{\theta \in \Theta} \dkl{\alpha_k Q_k +
    (1-\alpha_k)P_{\theta_k}}{P_{\theta}} \nonumber \\
  & = \argmax_{\theta \in \Theta} \left\{ \alpha_k \E_{Q_k}[\log p_{\theta}(X)]
    + (1-\alpha_k) \E_{\theta_k}[\log p_{\theta}(X)] \right\} \nonumber \\
  & = \argmax_{\theta \in \Theta} \left\{
    \alpha_k \theta^\top \E_{Q_k}[\suff(X)]
    + (1-\alpha_k) \theta^\top \nabla A(\theta_k) - A(\theta) \right\}.
    \label{eqn:ideal-ce}
\end{align}
The term $\E_{Q_k}[\suff(X)]$ is unknown in practice, so we use a
sampled estimate. For
$X_{k, 1}, \ldots, X_{k, N_k} \simiid P_{\theta_k}$, let $\thresh_k$ be the
$\tol$-quantile of $\obj(X_{k, 1}), \ldots, \obj(X_{k, N_k})$ and define
\begin{align}
  D_{k+1} \defeq \frac{1}{N_k} \sum_{i=1}^{N_k}
  \frac{q_k(X_{k, i})}{p_{\theta_k}(X_{k, i})} \suff(X_{k, i})
  = \frac{1}{N_k} \sum_{i=1}^{N_k}
  \frac{p_0(X_{k, i})}{p_{\theta_k}(X_{k, i})} \indic{\obj(X_{k, i}) \le \thresh_k}
  \suff(X_{k, i}).
    \label{eqn:is-ce}
\end{align}
Using the estimate $D_{k+1}$ in place of $\E_{Q_k}[\suff(X)]$ in the idealized
update~\eqref{eqn:ideal-ce}, we obtain Algorithm~\ref{alg:ce}. To select the
final importance sampling distribution from Algorithm~\ref{alg:ce}, we choose
$\theta_k$ with the lowest $\rho$-quantile of $f(X_{k, i})$. We observe that
this choice consistently improves performance over taking the last iterate or
Polyak averaging. Letting $\theta_{\rm ce}$ denote the parameters for the
importance sampling distribution learned by the cross-entropy method, we
sample $X_i \simiid P_{\theta_{\rm ce}}$ and use
$\what{p}_{N, \thresh} \defeq \frac{1}{N} \sum_{i=1}^N
\frac{p_0(X_i)}{p_{\theta_{\rm ce}}(X_i)} \indic{\obj(X_i) \le \thresh}$ as
our final importance-sampling estimator for $p_{\gamma}$.

In the context of our rare-event simulator, we use a combination of Beta and
Normal distributions for $P_\theta$. The sufficient statistics $\suff$ include
(i) the parameters of the generative model of behaviors that our imitation-learning schemes produce and (ii) the initial poses and velocities of other
vehicles, pedestrians, and obstacles in the simulation. Given a current
parameter $\theta$ and realization from the model distribution $P_\theta$, our
simulator then (i) sets the parameters of the generative model for vehicle
policies and draws policies from this model, and (ii) chooses random poses and
velocities for the simulation. Our simulator is one of the largest-scale
applications of cross-entropy methods.

\section{Simulation framework}
\label{sec:simulator}

Two key considerations in our risk-based framework influence design choices
for our simulation toolchain: (1) learning the base distribution $P_0$ of
nominal traffic behavior via data-driven modeling, and (2) testing the AV as a
\emph{whole system}. We now describe how our toolchain achieves these goals.

\subsection{Data-driven generative modeling}\label{sec:gail}

While our risk-based framework (cf.\ Section~\ref{sec:risk}) is a
concise, unambiguous measure of system safety, %
the rare-event probability $p_\gamma$ is only meaningful insofar as the base
distribution $P_0$ of road conditions and the behaviors of other (human)
drivers is estimable. Thus, to implement our risk-based framework, we first
learn a base distribution $P_0$ of nominal traffic behavior.  Using the
highway traffic dataset NGSim~\cite{Ngsim08}, we train policies of human drivers via imitation
learning~\cite{Russell98,RossBa10,RossGoBa11,HoEr16,BaramAnOrCasMa17}. Our
data consists of videos of highway traffic \cite{Ngsim08}, and our goal
is to create models that imitate human driving behavior even in scenarios distinct from those in the data. 
We employ an ensemble of generative adversarial imitation
learning (GAIL)~\cite{HoEr16} models to learn $P_0$. Our approach is motivated
by the observation that reducing an imitation-learning problem to supervised
learning---where we simply use expert data to predict actions given vehicle
states---suffers from poor performance in regions of the state space not
encountered in data~\cite{RossBa10,RossGoBa11}. Reinforcement-learning techniques
have been observed to improve generalization performance, as the imitation
agent is able to explore regions of the state space in simulation during
training that do not necessarily occur in the expert data traces.

Generically, GAIL is a minimax game between two functions: a discriminator
$D_{\phi}$ and a generator $G_{\xi}$ (with parameters $\phi$ and $\xi$
respectively). The discriminator takes in a state-action pair $(s,u)$ and
outputs the probability that the pair came from real data,
$\P(\text{real data})$. The generator takes in a state $s$ and outputs a
conditional distribution $G_{\xi}(s):=\P(u\mid s)$ of the action $u$ to take
given state $s$. In our context, $G_{\xi}(\cdot)$ is then the (learned) policy
of a human driver given environmental inputs $s$. Training the generator
weights $\xi$ occurs in a reinforcement-learning paradigm with reward
$-\log(1-D_{\phi}(s, G_{\xi}(s)))$. We use the model-based variant of GAIL
(MGAIL)~\cite{BaramAnOrCasMa17} which renders this reward fully differentiable
with respect to $\xi$ over a simulation rollout, allowing efficient model
training. GAIL has been validated by~\citet{KueflerMoWhKo17} to realistically
mimic human-like driving behavior from the NGSim dataset across multiple
metrics. These include the similarity of low-level actions (speeds,
accelerations, turn-rates, jerks, and time-to-collision), as well as
higher-level behaviors (lane change rate, collision rate, hard-brake rate,
etc). See Appendix \ref{sec:videos} for a reference to an example video of the learned model driving in a scenario alongside data traces from human drivers.

Our importance sampling and cross-entropy methods use not just a single
instance of model parameters $\xi$, but rather a distribution over them to
form a generative model of human driving behavior. To model this distribution,
we use a (multivariate normal) parametric bootstrap over a trained ensemble of
generators $\xi^i, \; i=1,\dots,m$. Our models $\xi^i$ are high-dimensional
($\xi\in \R^d, \; d > m$) as they characterize the weights of large neural
networks, so we employ the graphical lasso \cite{friedman2008sparse} to fit the inverse covariance
matrix for our ensemble. This approach to modeling uncertainty in
neural-network weights is similar to the bootstrap approach of
\citet{OsbandBlPrRo16}. Other approaches include using dropout for
inference~\cite{GalGh16} and variational methods \cite{Graves11,
  BlundellCoKaKoWi15, KingmaSaWe15}.

While several open source driving simulators have been
proposed~\cite{Dosovitskiy17,airsim2017fsr,deepdrive2}, our problem
formulation requires unique features to allow sampling from a
continuous distribution of driving policies for environmental agents. Conditional on each sample of model parameters $\xi$, the simulator constructs a (random) rollout of vehicle behaviors according to $G_\xi$. Unlike other existing simulators, ours is designed to efficiently execute and update these policies as new samples $\xi$ are drawn for each rollout.

\subsection{System architecture}
\label{sec:sys-arch}

The second key characteristic of our framework is that it enables black-box testing the AV as a \textit{whole system}. Flaws in complex systems routinely occur at poorly specified interfaces between components, as interactions between processes can induce unexpected behavior.  Consequently, solely testing subcomponents of an AV control pipeline separately is insufficient~\cite{cyphy18}. Moreover, it is increasingly common for manufacturers to utilize software and hardware artifacts for which they do not have any whitebox model \cite{heinecke2004automotive,cheah2016combining}. We provide a concise but extensible language-agnostic interface to our benchmark world model so that common AV sensors such as cameras and lidar can provide the necessary inputs to induce vehicle actuation commands. 

Our simulator is a distributed, modular framework, which is necessary to support the inclusion of new AV systems and updates to the environment-vehicle policies. A benefit of this
design is that simulation rollouts are simple to parallelize.
In particular, we allow instantiation
of multiple simulations simultaneously, without requiring that
each include the entire set of components. For example, a desktop may support only one instance of Unreal Engine but could be capable of simulating 10
physics simulations in parallel; it would be impossible to fully
utilize the compute resource with a monolithic executable wrapping all engines together. Our architecture enables instances of the components
to be distributed on heterogeneous GPU compute clusters while maintaining
the ability to perform meaningful analysis locally on commodity desktops. In Appendix~\ref{section:scenario}, we detail our scenario specification, which describes how Algorithm \ref{alg:ce} maps onto our distributed architecture.

\section{Experiments}
\label{sec:experiments}

In this section, we demonstrate our risk-based framework on a multi-agent
highway scenario.
As the rare-event probability of interest $p_{\gamma}$ gets smaller, the
cross-entropy method learns to sample more rare events compared to naive Monte Carlo sampling; we empirically observe that the cross-entropy method produces $2$-$20$
times as many rare events as its naive counterpart. Our findings hold across
different ego-vehicle policies, base distributions $P_0$, and scenarios.

To highlight the modularity of our simulator, we evaluate the rare-event
probability $p_{\gamma}$ on two different ego-vehicle policies. The first is
an instantiation of an imitation learning (non-vision) policy which uses lidar as its primary perceptual input. Secondly, we investigate a vision-based controller (vision policy), where the ego-vehicle
drives with an end-to-end highway autopilot network~\cite{bojarski2016end},
taking as input a rendered image from the simulator (and lidar observations)
and outputting actuation commands. See Appendix~\ref{section:architecture} for
a summary of network architectures used.

We consider a scenario consisting of six agents, five of which are considered
part of the environment. The environment vehicles' policies follow the
distribution learned in Section~\ref{sec:gail}. All vehicles are constrained
to start within a set of possible initial configurations consisting of pose
and velocity, and each vehicle has a goal of reaching the end of the
approximately 2 km stretch of road. Fig.~\ref{fig:i80scenario} shows one such
configuration of the scenario, along with rendered images from the
simulator. We create scene geometry based on surveyors' records and
photogrammetric reconstructions of satellite imagery of the portion of I-80 in
Emeryville, California where the traffic data was collected \cite{Ngsim08}.

\ifarxiv
\paragraph*{Simulation parameters}
\else
    \textbf{Simulation parameters}~~
\fi
We detail our postulated base distribution
$P_0$. Letting $m$ denote the number of vehicles, we consider the random tuple
$X = (S, T, W, V, \xi)$ as our simulation parameter where the pair
$(S, T) \in \R^{m \times 2}_+$ indicates the two-dimensional positioning of
each vehicle in their respective lanes (in meters), $W$ the
orientation of each vehicle (in degrees), and $V$ the initial velocity of each vehicle
(in meters per second). We use $\xi \in \R^{404}$ to denote the weights of the
last layer of the neural network trained to imitate human-like driving
behavior. Specifically, we set $S \sim 40\text{Beta}(2,2) + 80 $ with respect to the
starting point of the road, $T \sim 0.5\text{Beta}(2,2) -0.25$ with respect
to the lane's center, $W \sim 7.2\text{Beta}(2,2) -3.6$ with respect to
facing forward, and $V \sim 10\text{Beta}(2,2) + 10$. We assume
$\xi \sim \mathcal{N}(\mu_0, \Sigma_0)$, with the mean and covariance matrices
learned via the ensemble approach outlined in Section \ref{sec:gail}. The
neural network whose last layer is parameterized by $\xi$ describes the policy
of environment vehicles; it takes as input the state of the vehicle and lidar
observations of the surrounding environment (see Appendix \ref{section:architecture} for more details).  Throughout this section, we
define our measure of safety $f: \mc{X} \to \R$ as the minimum
time-to-collision (TTC) over the simulation rollout. We calculate TTC from the
center of mass of the ego vehicle; if the ego-vehicle's body crashes into
obstacles, we end the simulation before the TTC can further decrease (see Appendix \ref{section:scenario} for details).

\begin{figure}[!!!!!!t]
\begin{minipage}{0.49\columnwidth}
  \centering \subfigure[Ratio of number of rare events
  vs.
  threshold]{\includegraphics[width=0.9\textwidth]{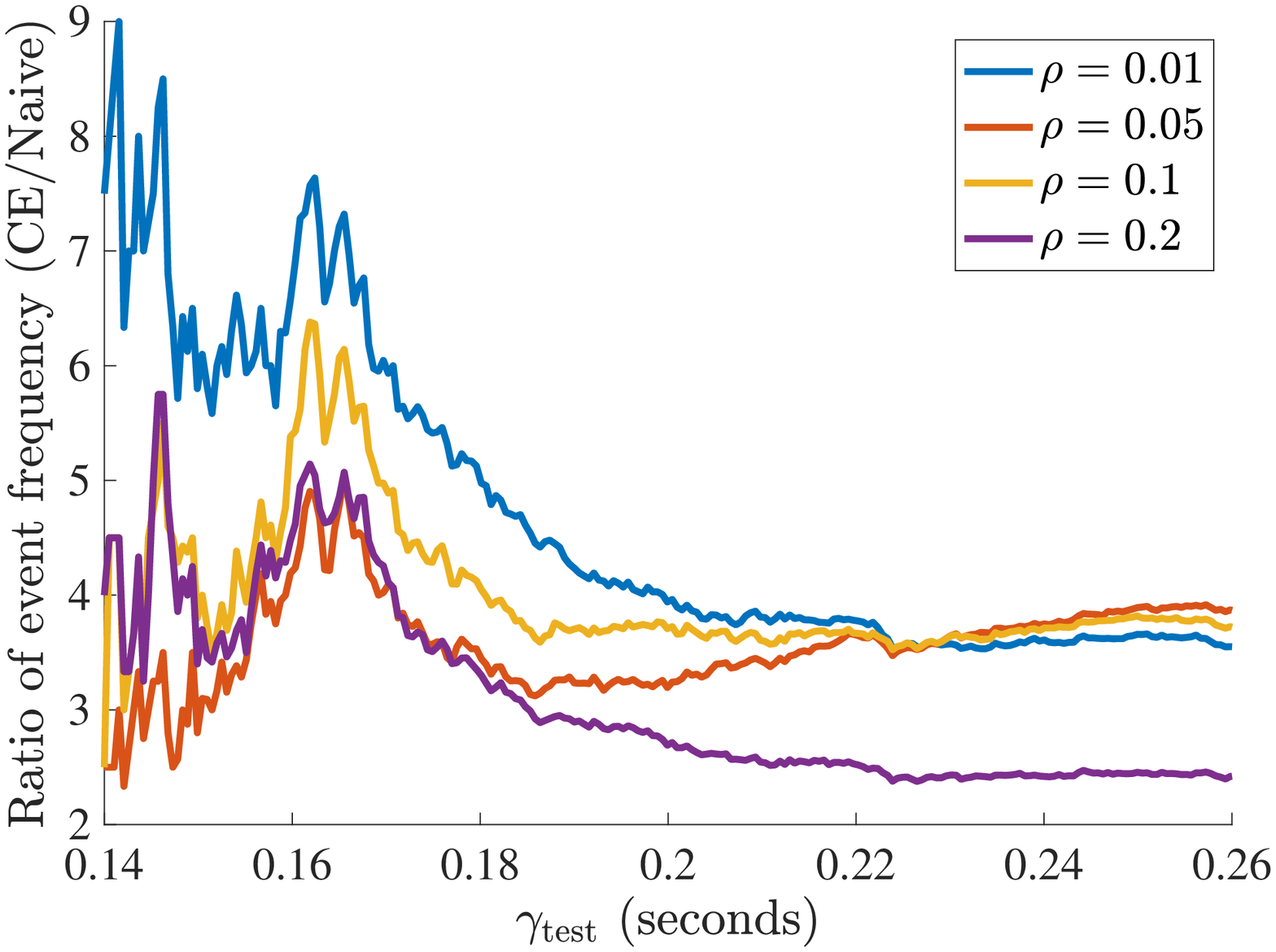}}%
\end{minipage}
\centering
\begin{minipage}{0.49\columnwidth}%
\centering
\subfigure[Ratio of variance vs. threshold]{\includegraphics[width=0.9\textwidth]{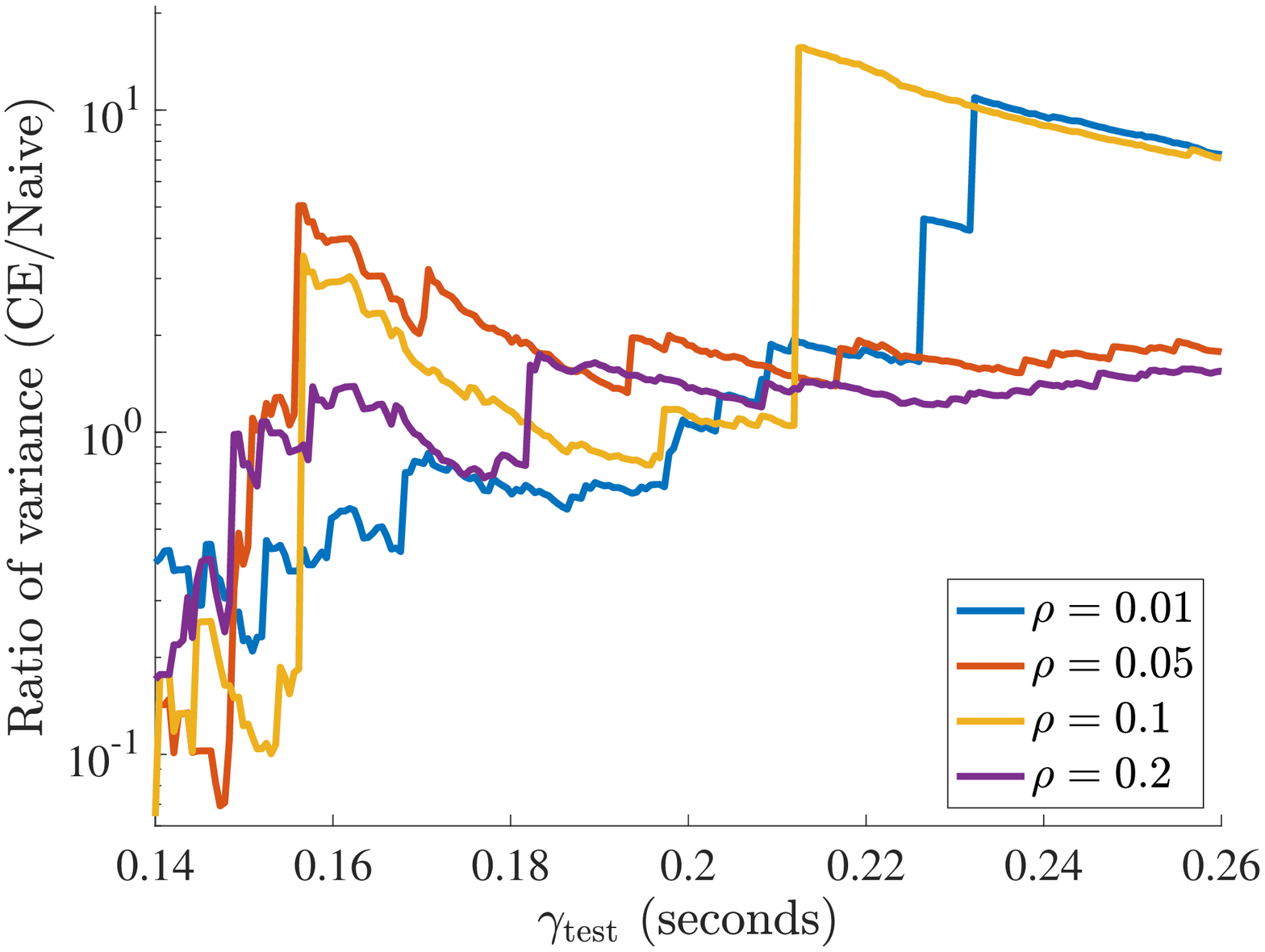}}%
\end{minipage}
\centering
\caption[]{\label{fig:rho} The ratio of $(a)$ number of rare events and $(b)$
  variance of estimator for $p_{\gamma}$ between cross-entropy method and
  naive MC sampling for the non-vision ego policy.  Rarity is inversely proportional to $\gamma$, and, as expected, we see the best performance for our method over naive MC at small $\gamma$. }
\end{figure}
\begin{figure}[!!!!!t]
\begin{center}
\footnotesize{
\begin{tabular}{c|lllll}
\multicolumn{1}{c|}{Search Algorithm} & \multicolumn{1}{c}{$\gamma_{\rm test}=0.14$} & \multicolumn{1}{c}{$\gamma_{\rm test}=0.15$} & \multicolumn{1}{c}{$\gamma_{\rm test}=0.19$} & \multicolumn{1}{c}{$\gamma_{\rm test}=0.20$} \\
\hline
  \hline
  Naive $1300$K    &   (12.4$\pm$3.1)e-6 &  (80.6$\pm$7.91)e-6  & (133$\pm$3.2)e-5   &  (186$\pm$3.79)e-5 \\ \hline 
Cross-entropy $100$K   &  (19.8$\pm$8.88)e-6&  (66.1 $\pm$ 15)e-6 &  (108$\pm$ 9.51)e-5 & (164 $\pm$ 14)e-5      \\
Naive $100$K   & (20$\pm$14.1)e-6 & (100$\pm$ 31.6)e-6& (132$\pm$11.5)e-5& (185$\pm$13.6)e-5    \\

\hline
\hline
\end{tabular}
}
\end{center}

\captionof{table}{\label{tab:pgamma_novision} Estimate of rare-event
  probability $p_{\gamma}$ (non-vision ego policy) with standard errors. For the cross-entropy method, we show results for the learned importance sampling distribution with $\tol = 0.01$.}
\end{figure}

\ifarxiv
    \paragraph*{Cross-entropy method}
\else
    \textbf{Cross-entropy method}~~ 
\fi
Throughout our experiments, we impose
constraints on the space of importance samplers (adversarial distributions)
for feasibility.  Numerical stability considerations predominantly drive our hyperparameter choices. For model
parameters $\xi$, we also constrain the search space to ensure that generative
models $G_\xi$ maintain reasonably realistic human-like policies (recall
Sec.~\ref{sec:gail}). For $S, T, W$, and $V$, we let $\{\mbox{Beta}(\alpha,
\beta): \alpha, \beta \in [1.5, 7] \}$ be the model space over which the
cross-entropy method searches, scaled and centered appropriately to match
the scale of the respective base distributions. We restrict the search
space of distributions over $\xi \in \R^{404}$ by searching over $\{
\normal(\mu, \Sigma_0): \linf{\mu - \mu_0} \le .01\}$, where
$(\mu_0,\Sigma_0)$ are the parameters of the base (bootstrap)
distribution. For our importance sampling distribution $P_{\theta}$, we use
products of the above marginal distributions. These
restrictions on the search space mitigate numerical instabilities in computing likelihood ratios within our
optimization routines, which is important for our high-dimensional problems.

We first illustrate the dependence of the cross-entropy method on its
hyperparameters. We choose to use a non-vision ego-vehicle policy as a test
bed for hyperparameter tuning, since this allows us to take advantage of the
fastest simulation speeds for our experiments. We focus on the effects (in
Algorithm~\ref{alg:ce}) of varying the most influential hyperparameter,
$\rho \in (0, 1]$, which is the quantile level determining the rarity
of the observations used to compute the importance sampler
$\theta_k$. Intuitively, as $\rho$ approaches 0, the cross-entropy method
learns importance samplers $P_{\theta}$ that up-weight unsafe regions of
$\mc{X}$ with lower $f(x)$, increasing the frequency of sampling rare events
(events with $\obj(X) \le \gamma$).  In order to avoid overfitting
$\theta_k$ as $\rho \to 0$, we need to increase $N_k$ as $\rho$
decreases. Our choice of $N_k$ is borne out of computational constraints as
it is the biggest factor that determines the run-time of the cross-entropy
method. Consistent with prior works~\cite{RubinsteinKr04, HuHu09}, we
observe empirically that $\rho \in [0.01, 0.2]$ is a good range for the values
of $N_k$ deemed feasible for our computational budget ($N_k = 1000 \sim
5000$). We fix the number of iterations at $K = 100$, number of samples
taken per iteration at $N_k = 5000$, step size for updates at $\alpha_k =
0.8$, and $\gamma = 0.14$. As we see below, we consistently observe that the
cross-entropy method learns to sample significantly more rare events,
despite the high-dimensional nature $(d \approx 500)$ of the problem.

To evaluate the learned parameters, we draw $n = 10^{5}$ samples from the
importance sampling distribution to form an estimate of $p_{\gamma}$.  In
Figure~\ref{fig:rho}, we vary $\rho$ and report the relative performance of
the cross-entropy method compared to naive Monte Carlo sampling. Even though
we set $\gamma=0.14$ in Algorithm \ref{alg:ce}, we evaluate the performance of
all models with respect to multiple threshold levels $\gamma_{\rm test}$. We
note that as $\rho$ approaches $0$, the cross-entropy method learns to
frequently sample increasingly rare events; the cross-entropy method yields
$3$-$10$ times as many dangerous scenarios, and achieves $2$-$16$ times
variance reduction depending on the threshold level $\gamma_{\rm test}$. In
Table~\ref{tab:pgamma_novision}, we contrast the estimates provided by naive
Monte Carlo and the importance sampling estimator provided by the
cross-entropy method with $\rho=0.01$; to form a baseline estimate, we run
naive Monte Carlo with $1.3 \cdot 10^{6}$ samples. For a given number of
samples, the cross-entropy method with $\rho=0.01$ provides more precise
estimates for the rare-event probability $p_{\gamma} \approx 10^{-5}$ over
naive Monte Carlo.

\begin{figure}[!!t]
\begin{minipage}{0.49\columnwidth}
  \centering \subfigure[Ratio of number of rare events
  vs.
  threshold]{\includegraphics[width=0.9\textwidth]{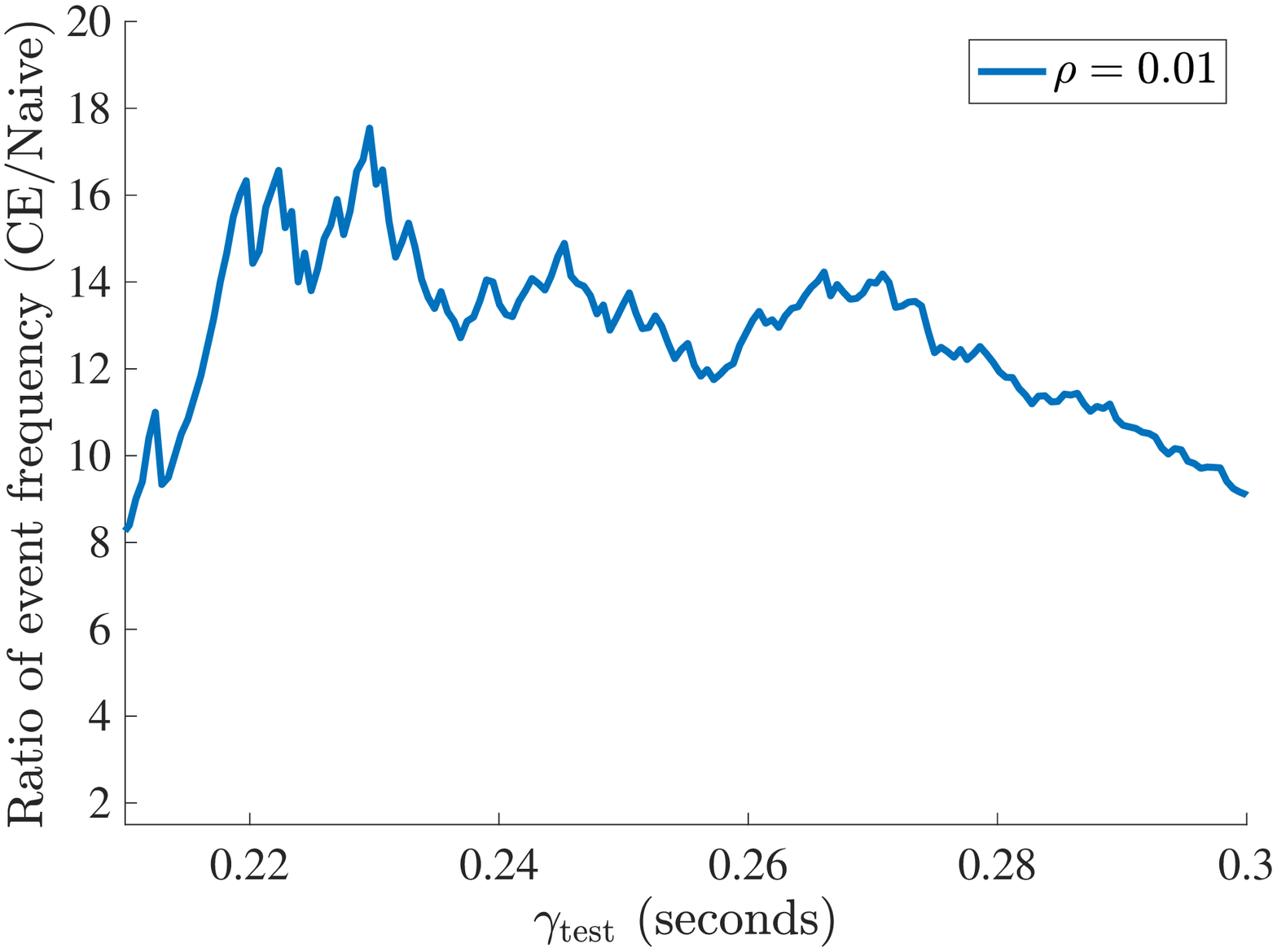}}%
\end{minipage}
\centering
\begin{minipage}{0.49\columnwidth}%
\centering
\subfigure[Ratio of variance vs. threshold]{\includegraphics[width=0.9\textwidth]{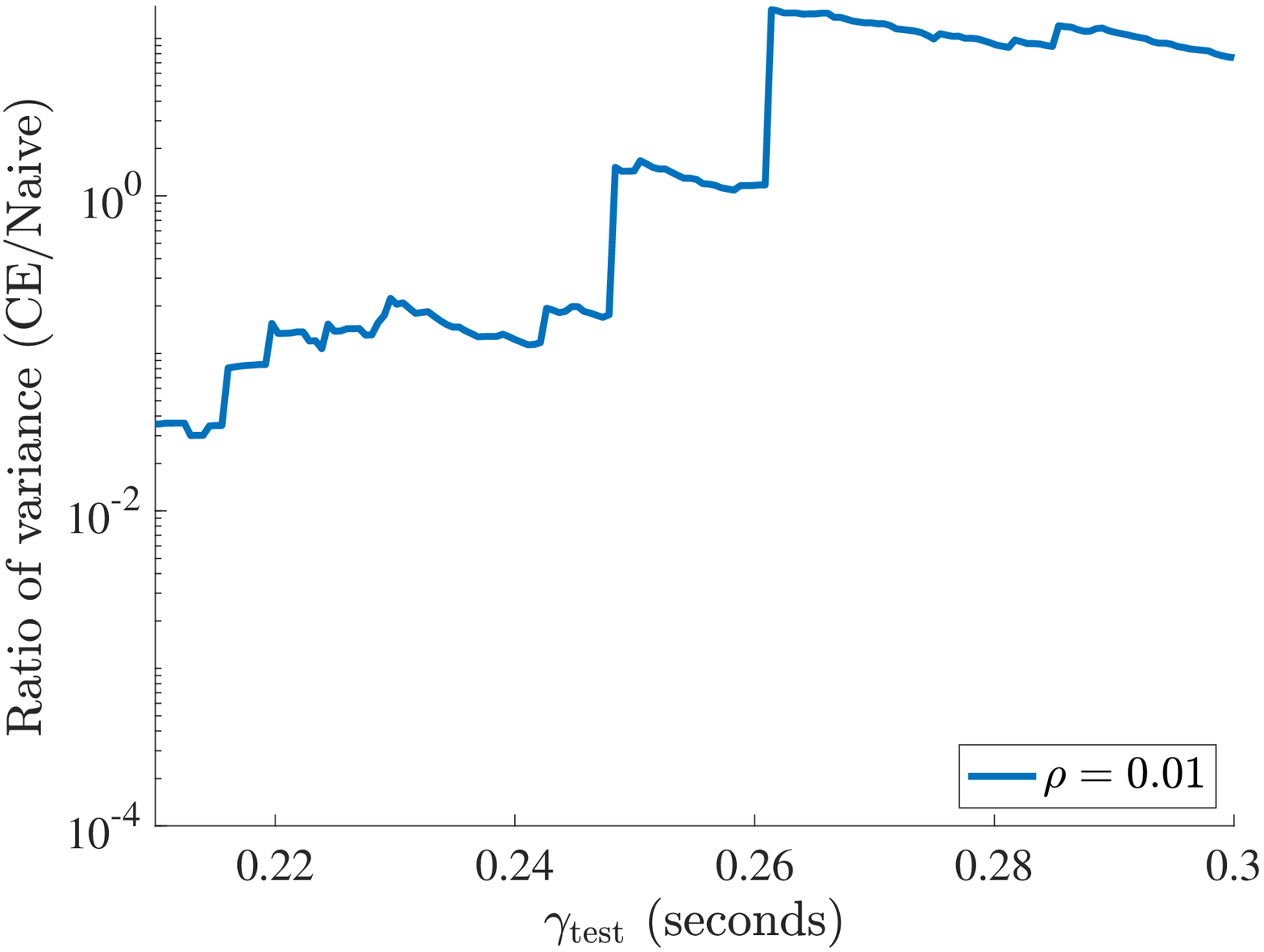}}%
\end{minipage}
\centering
\caption[]{\label{fig:vision} The ratio of $(a)$ number of rare events and
  $(b)$ variance of estimator for $p_{\gamma}$ between cross-entropy method
  and naive MC sampling for the vision-based ego policy.  }
\end{figure}

We now leverage the tuned hyperparameter ($\tol=0.01$) for our main
experiment: evaluating the probability of a dangerous event for the
vision-based ego policy. We find that the hyperparameters for the
cross-entropy method generalize, allowing us to produce good importance
samplers for a very different policy without further tuning. Based on our
computational budget (with our current implementation, vision-based
simulations run about 15 times slower than simulations with only non-vision
policies), we choose $K = 20$ and $N_k = 1000$ for the cross-entropy method to
learn a good importance sampling distribution for the vision-based policy
(although we also observe similar behavior for $N_k$ as small as $100$). In
Figure~\ref{fig:vision}, we illustrate again that the cross-entropy method
learns to sample dangerous scenarios more frequently
(Figure~\ref{fig:vision}a)---up to $18$ times that of naive Monte Carlo---and
produces importance sampling estimators with lower variance
(Figure~\ref{fig:vision}b). As a result, our estimator in
Table~\ref{tab:pgamma_vision} is better calibrated compared to that computed
from naive Monte Carlo.

\ifarxiv
    \paragraph*{Qualitative analysis}
\else
    \textbf{Qualitative analysis}~~
\fi
We provide a qualitative interpretation for the learned parameters of the
importance sampler. For initial velocities, angles, and positioning of
vehicles, the importance sampler shifts environmental vehicles to box in the
ego-vehicle and increases the speeds of trailing vehicles by $20\%$, making
accidents more frequent. We also observe that the learned distribution for
initial conditions have variance $50\%$ smaller than that of the base
distribution, implying concentration around adversarial conditions. Perturbing
the policy weights $\xi$ for GAIL increases the frequency of risky high-level
behaviors (lane-change rate, hard-brake rate, etc.). An interesting consequence of using our definition of TTC from the center of the ego vehicle (cf. Appendix \ref{section:scenario}) as a measure of safety is that dangerous
events $f(X) \le \gamma_{\rm test}$ (for small $\gamma_{\rm test}$) include
frequent sideswiping behavior, as such accidents result in smaller TTC values
than front- or rear-end collisions. See Appendix \ref{sec:videos} for a
reference to supplementary videos that exhibit the range of behavior across
many levels $\gamma_{\rm test}$. The modularity of our simulation framework
easily allows us to modify the safety objective to an alternative definition of TTC or even include more sophisticated
notions of safety, \eg~temporal-logic specifications or implementations of
responsibility-sensitive safety (RSS) \cite{shalev2017formal, roohi2018self}.

\begin{figure}[!!!!!t]
\begin{center}
\footnotesize{
\begin{tabular}{c|lllll}
\multicolumn{1}{c|}{Search Algorithm} & \multicolumn{1}{c}{$\gamma_{\rm test}=0.22$} & \multicolumn{1}{c}{$\gamma_{\rm test}=0.23$} & \multicolumn{1}{c}{$\gamma_{\rm test}=0.24$} & \multicolumn{1}{c}{$\gamma_{\rm test}=0.25$} \\
\hline
  \hline
Cross-entropy $50$K  &  (5.87$\pm$1.82)e-5 & (13.0$\pm$ 2.94)e-5  & (19.0 $\pm$ 3.14)e-5 & (4.52 $\pm$ 1.35)e-4      \\
Naive $50$K        & (11.3$\pm$4.60)e-5   & (20.6$\pm$6.22)e-5  & (43.2$\pm$9.00)e-5 & (6.75$\pm$1.13)e-4    \\

\hline
\hline
\end{tabular}
}
\end{center}

\captionof{table}{\label{tab:pgamma_vision} Estimate of rare-event
  probability $p_{\gamma}$ (non-vision ego policy) with standard errors. For the cross-entropy method, we show results for the learned importance sampling distribution with $\tol = 0.01$.}
\end{figure}

\section{Related work and conclusions}

Given the complexity of AV software and hardware components, it
is unlikely that any single method will serve as an oracle for
certification. Many existing tools are complementary to our risk-based framework. In this section, we compare
and contrast representative results in testing, verification, and simulation.
AV testing generally consists of three paradigms. The first, largely
attributable to regulatory efforts, uses a finite set of basic
competencies (\textit{e.g.}~the Euro NCAP Test Protocol
\cite{schram2013implementation}); while this methodology is successful in
designing safety features such as airbags and seat-belts, the non-adaptive
nature of static testing is less effective in complex software systems found
in AVs. Alternatively, real-world testing---deployment of vehicles with human
oversight---exposes the vehicle to a wider variety of unpredictable test
conditions. However, as we outlined above, these methods pose a danger to the
public and require prohibitive number of driving hours due to the rare nature
of accidents~\cite{KalraPa16}. Simulation-based falsification (in our context,
simply finding any crash) has also been successfully utilized
\cite{tuncali2016utilizing}; this approach does not maintain a link to the
likelihood of the occurrence of a particular event, which we believe to be key
in acting to prioritize and correct AV behavior.

Formal verification methods~\cite{KwiatkowskaNoPa11,Althoff2014,SeshiaSaSa15,
  apex_SAE16} have emerged as a candidate to reduce the intractability of
empirical validation. A verification procedure considers whether the system
can \emph{ever} violate a specification and returns either a proof that there
is no such execution or a
counterexample. Verification procedures require a white-box description of the system (although
it may be abstract), as well as a mathematically precise specification. Due to the impossibility of
certifying safety in \emph{all} scenarios, these
approaches~\cite{shalev2017formal} require further specifications that assign
blame in the case of a crash. Such assignment of blame is impossible to
completely characterize and relies on subjective notions of fault. Our
 risk-based framework allows one to circumvent this difficulty by only using a
measure of safety that does not assign blame (e.g. TTC) and replacing the specifications that assign blame
with a probabilistic notion of how likely the accident is. While this approach
requires a learned model of the world $P_0$---a highly nontrivial statistical
task in itself---the adaptive importance sampling techniques we employ can
still efficiently identify dangerous scenarios even when $P_0$ is not
completely accurate. Conceptually, we view verification and our framework as
complementary; they form powerful tools that can evaluate safety \emph{before}
deploying a fleet for real-world testing.

Even given a consistent and complete notion of blame, verification remains
highly intractable from a computational standpoint. Efficient algorithms only
exist for restricted classes of systems in the domain of AVs, and they are
fundamentally difficult to scale. Specifically, AVs---unlike previous
successful applications of verification methods to application domains such as
microprocessors~\cite{Baier2008}---include both continuous and discrete
dynamics.  This class of dynamics falls within the purview of hybrid
systems~\cite{lygeros2004lecture}, for which exhaustive verification is largely undecidable~\cite{HenzingerKPV98jcss}.

Verifying individual components of the perception pipeline, even as standalone systems, is a nascent, active area of research (see~\cite{AroraBhGeMa14,
  CohenShSh16,BartlettFoTe17} and many others). Current subsystem verification
techniques for deep neural
networks~\cite{HuangKwWaWu17,KatzBaDiJuKo17,TjengTe17} do not scale to
state-of-the-art models and largely investigate the robustness of the network
with respect to small perturbations of a single sample. There are two key
assumptions in these works; the label of the input is unchanged within the
radius of allowable perturbations, and the resulting expansion of the test set
covers a meaningful portion of possible inputs to the network. Unfortunately,
for realistic cases in AVs it is likely that perturbations to the state of the
world which in turn generates an image \textit{should} change the
label. Furthermore, the combinatorial nature of scenario configurations casts
serious doubt on any claims of coverage.

In our risk-based framework, we replace the complex system specifications
required for formal verification methods with a model $P_0$ that we learn via
imitation-learning techniques. Generative adversarial imitation learning
(GAIL) was first introduced by~\citet{HoEr16} as a way to directly learn
policies from data and has since been applied to model human driving behavior
by~\citet{KueflerMoWhKo17}. Model-based GAIL (MGAIL) is the specific variant
of GAIL that we employ; introduced by \citet{BaramAnOrCasMa17}, MGAIL's
generative model is fully differentiable, allowing efficient model training
with standard stochastic approximation methods.

The cross-entropy method was introduced by~\citet{Rubinstein01} and has
attracted interest in many rare-event simulation
scenarios~\cite{RubinsteinKr04, KroeseRuGl13}.  More broadly, it can be
thought of as a model-based optimization method~\cite{HuHu09,HuHu11,
  HuHuCh12,Zabinsky13,HuZhFa14,ZhouHu14}.  With respect to assessing safety of
AVs, the cross-entropy method has recently been applied in simple
lane-changing and car-following scenarios in two dimensions~\cite{Zhao16,
  ZhaoHuPeLaLe18}. Our work significantly extends these works by implementing a
photo-realistic simulator that can assess the deep-learning based perception
pipeline along with the control framework. We leave the development of rare-event
simulation methods that scale better with dimension as a future work.

To summarize, a fundamental tradeoff emerges when comparing the requirements
of our risk-based framework to other testing paradigms, such as real-world
testing or formal verification. Real-world testing endangers the public but is
still in some sense a gold standard. Verified subsystems provide evidence that 
the AV should drive safely even if the estimated distribution
shifts, but verification techniques are limited by computational intractability as well as the need for
both white-box models and the completeness of
specifications that assign blame (\textit{e.g.}~\cite{shalev2017formal}). 
In turn, our risk-based framework is most useful when the base distribution $P_0$
 is accurate, but even when $P_0$ is misspecified, our adaptive importance sampling 
 techniques can still efficiently identify dangerous scenarios, especially those that may be missed 
 by verification methods assigning blame. Our framework offers significant speedups over real-world testing
 and allows efficient evaluation of black-box AV input/output behavior, providing a powerful tool to 
 aid in the design of safe AVs.

\subsubsection*{Acknowledgments}
MOK was partially supported by a National Science Foundation Graduate Research Fellowship. AS was partially supported by a Stanford Graduate Fellowship and a Fannie \& John Hertz Foundation Fellowship. HN was partially supported by a Samsung Fellowship and the SAIL-Toyota Center for AI Research. JD was partially supported by the SAIL-Toyota Center for AI Research and National Science Foundation award NSF-CAREER-1553086. %

\setlength{\bibsep}{3pt}
\bibliography{bib,mok}

\begin{thebibliography}{56}
\providecommand{\natexlab}[1]{#1}
\providecommand{\url}[1]{\texttt{#1}}
\expandafter\ifx\csname urlstyle\endcsname\relax
  \providecommand{\doi}[1]{doi: #1}\else
  \providecommand{\doi}{doi: \begingroup \urlstyle{rm}\Url}\fi

\bibitem[Abbas et~al.(2018)Abbas, O'Kelly, Rodionova, and Mangharam]{cyphy18}
H.~Abbas, M.~O'Kelly, A.~Rodionova, and R.~Mangharam.
\newblock Safe at any speed: A simulation-based test harness for autonomous
  vehicles.
\newblock LNCS. Springer, 2018.

\bibitem[Althoff and Dolan(2014)]{Althoff2014}
M.~Althoff and J.~Dolan.
\newblock Online verification of automated road vehicles using reachability
  analysis.
\newblock \emph{Robotics, IEEE Transactions on}, 30\penalty0 (4):\penalty0
  903--918, Aug 2014.
\newblock ISSN 1552-3098.
\newblock \doi{10.1109/TRO.2014.2312453}.

\bibitem[Arora et~al.(2014)Arora, Bhaskara, Ge, and Ma]{AroraBhGeMa14}
S.~Arora, A.~Bhaskara, R.~Ge, and T.~Ma.
\newblock Provable bounds for learning some deep representations.
\newblock In \emph{International Conference on Machine Learning}, pages
  584--592. ~, 2014.

\bibitem[Asmussen and Glynn(2007)]{AsmussenGl07}
S.~Asmussen and P.~W. Glynn.
\newblock \emph{Stochastic Simulation: Algorithms and Analysis}.
\newblock Springer, 2007.

\bibitem[Baier et~al.(2008)Baier, Katoen, et~al.]{Baier2008}
C.~Baier, J.-P. Katoen, et~al.
\newblock \emph{Principles of model checking}, volume 26202649.
\newblock MIT press Cambridge, 2008.

\bibitem[Baram et~al.(2017)Baram, Anschel, Caspi, and Mannor]{BaramAnOrCasMa17}
N.~Baram, O.~Anschel, I.~Caspi, and S.~Mannor.
\newblock End-to-end differentiable adversarial imitation learning.
\newblock In \emph{International Conference on Machine Learning}, pages
  390--399, 2017.

\bibitem[Bartlett et~al.(2017)Bartlett, Foster, and Telgarsky]{BartlettFoTe17}
P.~L. Bartlett, D.~J. Foster, and M.~J. Telgarsky.
\newblock Spectrally-normalized margin bounds for neural networks.
\newblock In \emph{Advances in Neural Information Processing Systems}, pages
  6241--6250. ~, 2017.

\bibitem[Blundell et~al.(2015)Blundell, Cornebise, Kavukcuoglu, and
  Wierstra]{BlundellCoKaKoWi15}
C.~Blundell, J.~Cornebise, K.~Kavukcuoglu, and D.~Wierstra.
\newblock Weight uncertainty in neural networks.
\newblock \emph{arXiv preprint arXiv:1505.05424}, 2015.

\bibitem[Bojarski et~al.(2016)Bojarski, Del~Testa, Dworakowski, Firner, Flepp,
  Goyal, Jackel, Monfort, Muller, Zhang, et~al.]{bojarski2016end}
M.~Bojarski, D.~Del~Testa, D.~Dworakowski, B.~Firner, B.~Flepp, P.~Goyal, L.~D.
  Jackel, M.~Monfort, U.~Muller, J.~Zhang, et~al.
\newblock End to end learning for self-driving cars.
\newblock \emph{arXiv preprint arXiv:1604.07316}, 2016.

\bibitem[Bonnefon et~al.(2016)Bonnefon, Shariff, and Rahwan]{Bonnefon1573}
J.-F. Bonnefon, A.~Shariff, and I.~Rahwan.
\newblock The social dilemma of autonomous vehicles.
\newblock \emph{Science}, 352\penalty0 (6293):\penalty0 1573--1576, 2016.
\newblock ISSN 0036-8075.
\newblock \doi{10.1126/science.aaf2654}.
\newblock URL \url{http://science.sciencemag.org/content/352/6293/1573}.

\bibitem[Bucklew(2013)]{Bucklew13}
J.~Bucklew.
\newblock \emph{Introduction to rare event simulation}.
\newblock Springer Science \& Business Media, 2013.

\bibitem[Cheah et~al.(2016)Cheah, Shaikh, Bryans, and
  Nguyen]{cheah2016combining}
M.~Cheah, S.~A. Shaikh, J.~Bryans, and H.~N. Nguyen.
\newblock Combining third party components securely in automotive systems.
\newblock In \emph{IFIP International Conference on Information Security Theory
  and Practice}, pages 262--269. Springer, 2016.

\bibitem[Cohen et~al.(2016)Cohen, Sharir, and Shashua]{CohenShSh16}
N.~Cohen, O.~Sharir, and A.~Shashua.
\newblock On the expressive power of deep learning: A tensor analysis.
\newblock In \emph{Conference on Learning Theory}, pages 698--728. ~, 2016.

\bibitem[Dosovitskiy et~al.(2017)Dosovitskiy, Ros, Codevilla, Lopez, and
  Koltun]{Dosovitskiy17}
A.~Dosovitskiy, G.~Ros, F.~Codevilla, A.~Lopez, and V.~Koltun.
\newblock {CARLA}: {An} open urban driving simulator.
\newblock In \emph{Proceedings of the 1st Annual Conference on Robot Learning},
  pages 1--16, 2017.

\bibitem[Friedman et~al.(2008)Friedman, Hastie, and
  Tibshirani]{friedman2008sparse}
J.~Friedman, T.~Hastie, and R.~Tibshirani.
\newblock Sparse inverse covariance estimation with the graphical lasso.
\newblock \emph{Biostatistics}, 9\penalty0 (3):\penalty0 432--441, 2008.

\bibitem[Gal and Ghahramani(2016)]{GalGh16}
Y.~Gal and Z.~Ghahramani.
\newblock Dropout as a bayesian approximation: Representing model uncertainty
  in deep learning.
\newblock In \emph{International Conference on Machine Learningearning}, pages
  1050--1059, 2016.

\bibitem[Games(2015)]{games2015unreal}
E.~Games.
\newblock Unreal engine 4 documentation.
\newblock \emph{URL https://docs. unrealengine. com/latest/INT/index. html},
  2015.

\bibitem[Graves(2011)]{Graves11}
A.~Graves.
\newblock Practical variational inference for neural networks.
\newblock In \emph{Advances in Neural Information Processing Systems}, pages
  2348--2356, 2011.

\bibitem[Heinecke et~al.(2004)Heinecke, Schnelle, Fennel, Bortolazzi, Lundh,
  Leflour, Mat{\'e}, Nishikawa, and Scharnhorst]{heinecke2004automotive}
H.~Heinecke, K.-P. Schnelle, H.~Fennel, J.~Bortolazzi, L.~Lundh, J.~Leflour,
  J.-L. Mat{\'e}, K.~Nishikawa, and T.~Scharnhorst.
\newblock Automotive open system architecture-an industry-wide initiative to
  manage the complexity of emerging automotive e/e-architectures.
\newblock Technical report, SAE Technical Paper, 2004.

\bibitem[Henzinger et~al.(1998)Henzinger, Kopke, Puri, and
  Varaiya]{HenzingerKPV98jcss}
T.~A. Henzinger, P.~W. Kopke, A.~Puri, and P.~Varaiya.
\newblock What's decidable about hybrid automata?
\newblock \emph{J. Comput. Syst. Sci.}, 57\penalty0 (1):\penalty0 94--124,
  1998.

\bibitem[Hintjens(2013)]{Hintjens13}
P.~Hintjens.
\newblock \emph{ZeroMQ: messaging for many applications}.
\newblock " O'Reilly Media, Inc.", 2013.

\bibitem[Ho and Ermon(2016)]{HoEr16}
J.~Ho and S.~Ermon.
\newblock Generative adversarial imitation learning.
\newblock In \emph{Advances in Neural Information Processing Systems}, pages
  4565--4573, 2016.

\bibitem[Homem-de Mello(2007)]{HomemDeMello07}
T.~Homem-de Mello.
\newblock A study on the cross-entropy method for rare-event probability
  estimation.
\newblock \emph{INFORMS Journal on Computing}, 19\penalty0 (3):\penalty0
  381--394, 2007.

\bibitem[Hu and Hu(2009)]{HuHu09}
J.~Hu and P.~Hu.
\newblock On the performance of the cross-entropy method.
\newblock In \emph{Simulation Conference (WSC), Proceedings of the 2009
  Winter}, pages 459--468. IEEE, 2009.

\bibitem[Hu and Hu(2011)]{HuHu11}
J.~Hu and P.~Hu.
\newblock Annealing adaptive search, cross-entropy, and stochastic
  approximation in global optimization.
\newblock \emph{Naval Research Logistics (NRL)}, 58\penalty0 (5):\penalty0
  457--477, 2011.

\bibitem[Hu et~al.(2012)Hu, Hu, and Chang]{HuHuCh12}
J.~Hu, P.~Hu, and H.~S. Chang.
\newblock A stochastic approximation framework for a class of randomized
  optimization algorithms.
\newblock \emph{IEEE Transactions on Automatic Control}, 57\penalty0
  (1):\penalty0 165--178, 2012.

\bibitem[Hu et~al.(2014)Hu, Zhou, and Fan]{HuZhFa14}
J.~Hu, E.~Zhou, and Q.~Fan.
\newblock Model-based annealing random search with stochastic averaging.
\newblock \emph{ACM Transactions on Modeling and Computer Simulation (TOMACS)},
  24\penalty0 (4):\penalty0 21, 2014.

\bibitem[Huang et~al.(2017)Huang, Kwiatkowska, Wang, and Wu]{HuangKwWaWu17}
X.~Huang, M.~Kwiatkowska, S.~Wang, and M.~Wu.
\newblock Safety verification of deep neural networks.
\newblock In \emph{International Conference on Computer Aided Verification},
  pages 3--29. Springer, 2017.

\bibitem[Kalra and Paddock(2016)]{KalraPa16}
N.~Kalra and S.~M. Paddock.
\newblock Driving to safety: How many miles of driving would it take to
  demonstrate autonomous vehicle reliability?
\newblock \emph{Transportation Research Part A: Policy and Practice},
  94:\penalty0 182--193, 2016.

\bibitem[Katz et~al.(2017)Katz, Barrett, Dill, Julian, and
  Kochenderfer]{KatzBaDiJuKo17}
G.~Katz, C.~Barrett, D.~Dill, K.~Julian, and M.~Kochenderfer.
\newblock Reluplex: An efficient smt solver for verifying deep neural networks.
\newblock \emph{arXiv:1702.01135 [cs.AI]}, 1:\penalty0 1, 2017.

\bibitem[Kingma et~al.(2015)Kingma, Salimans, and Welling]{KingmaSaWe15}
D.~P. Kingma, T.~Salimans, and M.~Welling.
\newblock Variational dropout and the local reparameterization trick.
\newblock In \emph{Advances in Neural Information Processing Systems}, pages
  2575--2583, 2015.

\bibitem[Kroese et~al.(2013)Kroese, Rubinstein, and Glynn]{KroeseRuGl13}
D.~P. Kroese, R.~Y. Rubinstein, and P.~W. Glynn.
\newblock The cross-entropy method for estimation.
\newblock \emph{Handbook of Statistics: Machine Learning: Theory and
  Applications}, 31:\penalty0 19--34, 2013.

\bibitem[Kuefler et~al.(2017)Kuefler, Morton, Wheeler, and
  Kochenderfer]{KueflerMoWhKo17}
A.~Kuefler, J.~Morton, T.~Wheeler, and M.~Kochenderfer.
\newblock Imitating driver behavior with generative adversarial networks.
\newblock In \emph{Intelligent Vehicles Symposium (IV), 2017 IEEE}, pages
  204--211. IEEE, 2017.

\bibitem[Kwiatkowska et~al.(2011)Kwiatkowska, Norman, and
  Parker]{KwiatkowskaNoPa11}
M.~Kwiatkowska, G.~Norman, and D.~Parker.
\newblock Prism 4.0: Verification of probabilistic real-time systems.
\newblock In \emph{International conference on computer aided verification},
  pages 585--591. Springer, 2011.

\bibitem[Lygeros(2004)]{lygeros2004lecture}
J.~Lygeros.
\newblock Lecture notes on hybrid systems.
\newblock In \emph{Notes for an ENSIETA workshop}, 2004.

\bibitem[of~Transportation~FHWA(2008)]{Ngsim08}
U.~D. of~Transportation~FHWA.
\newblock Ngsim -- next generation simulation, 2008.

\bibitem[O'Kelly et~al.(2016)O'Kelly, Abbas, Gao, Shiraishi, Kato, and
  Mangharam]{apex_SAE16}
M.~O'Kelly, H.~Abbas, S.~Gao, S.~Shiraishi, S.~Kato, and R.~Mangharam.
\newblock Apex: Autonomous vehicle plan verification and execution.
\newblock volume~1, Apr 2016.

\bibitem[Osband et~al.(2016)Osband, Blundell, Pritzel, and
  Van~Roy]{OsbandBlPrRo16}
I.~Osband, C.~Blundell, A.~Pritzel, and B.~Van~Roy.
\newblock Deep exploration via bootstrapped dqn.
\newblock In \emph{Advances in neural information processing systems}, pages
  4026--4034, 2016.

\bibitem[Quiter and Ernst(2018)]{deepdrive2}
C.~Quiter and M.~Ernst.
\newblock Deepdrive.
\newblock \url{https://github.com/deepdrive/deepdrive}, 2018.

\bibitem[Roohi et~al.(2018)Roohi, Kaur, Weimer, Sokolsky, and
  Lee]{roohi2018self}
N.~Roohi, R.~Kaur, J.~Weimer, O.~Sokolsky, and I.~Lee.
\newblock Self-driving vehicle verification towards a benchmark.
\newblock \emph{arXiv preprint arXiv:1806.08810}, 2018.

\bibitem[Ross and Bagnell(2010)]{RossBa10}
S.~Ross and D.~Bagnell.
\newblock Efficient reductions for imitation learning.
\newblock In \emph{Proceedings of the thirteenth international conference on
  artificial intelligence and statistics}, pages 661--668, 2010.

\bibitem[Ross et~al.(2011)Ross, Gordon, and Bagnell]{RossGoBa11}
S.~Ross, G.~Gordon, and D.~Bagnell.
\newblock A reduction of imitation learning and structured prediction to
  no-regret online learning.
\newblock In \emph{Proceedings of the fourteenth international conference on
  artificial intelligence and statistics}, pages 627--635, 2011.

\bibitem[Rubinstein(2001)]{Rubinstein01}
R.~Y. Rubinstein.
\newblock Combinatorial optimization, cross-entropy, ants and rare events.
\newblock In \emph{Stochastic optimization: algorithms and applications}, pages
  303--363. Springer, 2001.

\bibitem[Rubinstein and Kroese(2004)]{RubinsteinKr04}
R.~Y. Rubinstein and D.~P. Kroese.
\newblock \emph{The cross-entropy method: A unified approach to Monte Carlo
  simulation, randomized optimization and machine learning}.
\newblock Information Science \& Statistics, Springer Verlag, NY, 2004.

\bibitem[Russell(1998)]{Russell98}
S.~Russell.
\newblock Learning agents for uncertain environments.
\newblock In \emph{Proceedings of the eleventh annual conference on
  Computational learning theory}, pages 101--103. ACM, 1998.

\bibitem[Schram et~al.(2013)Schram, Williams, and van
  Ratingen]{schram2013implementation}
R.~Schram, A.~Williams, and M.~van Ratingen.
\newblock Implementation of autonomous emergency braking (aeb), the next step
  in euro ncap’s safety assessment.
\newblock \emph{ESV, Seoul}, 2013.

\bibitem[Seshia et~al.(2015)Seshia, Sadigh, and Sastry]{SeshiaSaSa15}
S.~A. Seshia, D.~Sadigh, and S.~S. Sastry.
\newblock Formal methods for semi-autonomous driving.
\newblock In \emph{Proceedings of the 52nd Annual Design Automation
  Conference}, page 148. ACM, 2015.

\bibitem[Shah et~al.(2017)Shah, Dey, Lovett, and Kapoor]{airsim2017fsr}
S.~Shah, D.~Dey, C.~Lovett, and A.~Kapoor.
\newblock Airsim: High-fidelity visual and physical simulation for autonomous
  vehicles.
\newblock In \emph{Field and Service Robotics}, 2017.
\newblock URL \url{https://arxiv.org/abs/1705.05065}.

\bibitem[Shalev-Shwartz et~al.(2017)Shalev-Shwartz, Shammah, and
  Shashua]{shalev2017formal}
S.~Shalev-Shwartz, S.~Shammah, and A.~Shashua.
\newblock On a formal model of safe and scalable self-driving cars.
\newblock \emph{arXiv preprint arXiv:1708.06374}, 2017.

\bibitem[Tjeng and Tedrake(2017)]{TjengTe17}
V.~Tjeng and R.~Tedrake.
\newblock Verifying neural networks with mixed integer programming.
\newblock \emph{arXiv:1711.07356 [cs.LG]}, 2017.

\bibitem[Tuncali et~al.(2016)Tuncali, Pavlic, and
  Fainekos]{tuncali2016utilizing}
C.~E. Tuncali, T.~P. Pavlic, and G.~Fainekos.
\newblock Utilizing s-taliro as an automatic test generation framework for
  autonomous vehicles.
\newblock In \emph{Intelligent Transportation Systems (ITSC), 2016 IEEE 19th
  International Conference on}, pages 1470--1475. IEEE, 2016.

\bibitem[Vogel(2003)]{vogel2003comparison}
K.~Vogel.
\newblock A comparison of headway and time to collision as safety indicators.
\newblock \emph{Accident analysis \& prevention}, 35\penalty0 (3):\penalty0
  427--433, 2003.

\bibitem[Zabinsky(2013)]{Zabinsky13}
Z.~B. Zabinsky.
\newblock \emph{Stochastic adaptive search for global optimization}, volume~72.
\newblock Springer Science \& Business Media, 2013.

\bibitem[Zhao(2016)]{Zhao16}
D.~Zhao.
\newblock \emph{Accelerated Evaluation of Automated Vehicles.}
\newblock {Ph.D.} thesis, Department of Mechanical Engineering, University of
  Michigan, 2016.

\bibitem[Zhao et~al.(2018)Zhao, Huang, Peng, Lam, and LeBlanc]{ZhaoHuPeLaLe18}
D.~Zhao, X.~Huang, H.~Peng, H.~Lam, and D.~J. LeBlanc.
\newblock Accelerated evaluation of automated vehicles in car-following
  maneuvers.
\newblock \emph{IEEE Transactions on Intelligent Transportation Systems},
  19\penalty0 (3):\penalty0 733--744, 2018.

\bibitem[Zhou and Hu(2014)]{ZhouHu14}
E.~Zhou and J.~Hu.
\newblock Gradient-based adaptive stochastic search for non-differentiable
  optimization.
\newblock \emph{IEEE Transactions on Automatic Control}, 59\penalty0
  (7):\penalty0 1818--1832, 2014.

\end{thebibliography}
\bibliographystyle{abbrvnat}

\newpage

\appendix

\section{Scenario specification}
\label{section:scenario}
A scenario specification consists of a scenario
description and outputs both $p_{\gamma}$~\eqref{eqn:rare-prob}, the accident
rate, and a dataset consisting of initial conditions and the minimum time to
collision, our continuous objective safety measure. Concretely, a scenario
description includes
\begin{itemize}
\item a set of possible initial conditions, e.g. a range of velocities and
  poses for each agent
\item a safety measure specification for the ego agent,
\item a generative model of environment policies, an ego vehicle model,
\item a world geometry model, \eg a textured mesh of the static scene in
  which the scenario is to take place.
\end{itemize}
Given the scenario description, the search module creates physics and
rendering engine worker instances, and Algorithm \ref{alg:ce} then adaptively searches through many perturbations of conditions in the scenario, which we call scenario realizations. A set of scenario realizations
may be mapped to multiple physics, rendering, and agent instantiations,
evaluated in parallel, and reduced by a sink node which reports a measure of
each scenarios performance relative to the specification.

In our implementation the safety measure is minimum time-to-collision (TTC). TTC is defined as the time it would take for two vehicles to intercept one another given that they each maintain their current heading and velocity \cite{vogel2003comparison}. The TTC between the ego-vehicle and vehicle $i$ is given by
\begin{equation}
TTC_i(t) = -\frac{r_i(t)}{\dot r_i(t)},
\end{equation}
where $r_i$ is the distance between the ego vehicle and vehicle $i$, and $\dot r_i$ the time derivative of this distance (which is simply computed by projecting the relative velocity of vehicle $i$ onto the vector between the vehicles' poses).

In this paper, vehicles are described as oriented rectangles in the 2D plane. Since we are interested in the time it would take for the ego-vehicle to intersect the polygonal boundary of another vehicle on the road, we utilize a finite set of range and range measurements in order to approximate the TTC metric. For a given configuration of vehicles, we compute $N$ uniformly spaced angles $\theta_1,\dots, \theta_N$ in the range $[0, 2\pi]$ with respect to the ego vehicle's orientation and cast rays outward from the center of the ego vehicle. For each direction we compute the distance which a ray could travel before intersecting one of the $M$ other vehicles in the environment. These form $N$ range measurements $s_1,\dots, s_N$. Further, for each ray $s_i$, we determine which vehicle (if any) that ray hit; projecting the relative velocity of this vehicle with respect to ego vehicle gives the range-rate measurement $\dot s_i$. Finally, we approximate the minimum TTC for a given simulation rollout $X$ of length $T$ discrete time steps by:
\begin{equation*}
f(X):= \min_{t=0,\dots,T} \left ( \min_{i=1,\dots, N} \frac{-s_i(t)}{\dot s_i(t)} \right ) 
\end{equation*}
Note that this measure can approximate the true TTC arbitrarily well via choice of $N$ and the discretization of time used by the simulator. Furthermore, note that our definition of TTC is with respect to the \emph{center} of the ego vehicle touching the \emph{boundary} of another vehicle. Crashing, on the other hand, is defined in our simulation as the intersection of boundaries of two vehicles. Thus, TTC values we evaluate in our simulation are nonzero even during crashes, since the center of the ego vehicle has not yet collided with the boundary of another vehicle.

\section{Network architectures}
\label{section:architecture}
\begin{figure}[!!t]
\centering
\includegraphics[width=.54\textwidth]{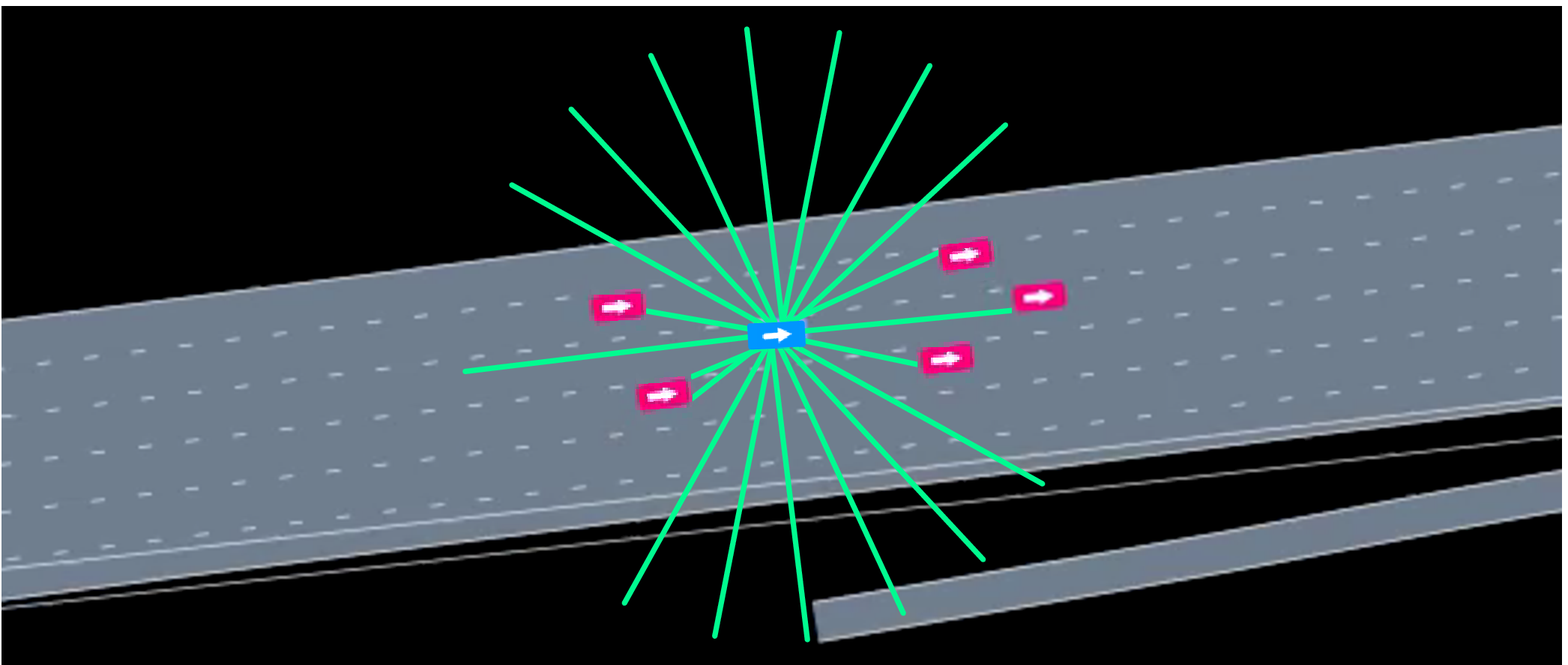}
\caption{Depiction of lidar sensor input used for GAIL models.}
\label{fig:lidar}
\end{figure}

The MGAIL generator model we use takes the same inputs as that of \citet{KueflerMoWhKo17}---the dynamical states of the vehicle as well as virtual lidar beam reflections. Specifically, we take as inputs: geometric parameters (vehicle length/width), dynamical states (vehicle speed, lateral and angular offsets with respect to the center and heading of the lane, distance to left and right lane boundaries, and local lane curvature), three indicators for collision, road departure, and traveling in reverse, and lidar sensor observations (ranges and range-rates of 20 lidar beams) as depicted in Figure \ref{fig:lidar}. The generator has two hidden layers of 200 and 100 neurons. The output consists of the mean and variance of normal distributions for throttle and steering commands; we then sample from these distributions to draw a given vehicle's action. The discriminator shares the same size for hidden layers. The forward model used to allow fully-differentiable training first encodes both the state and action through a 150 neuron layer and also adds a GRU layer to the state encoding. A Hadamard product of the results creates a joint embedding which is put through three hidden layers each of 150 neurons. The output is a prediction of the next state.

The end-to-end highway autopilot model is a direct implementation of \citet{bojarski2016end} via the code found at the link \url{https://github.com/sullychen/autopilot-tensorflow}. In our implementation of the vision-based policy, this highway autopilot model uses rendered images to produce steering commands. Lidar inputs are used to generate throttle commands using the same network as the non-vision policy. %
\section{Supplementary videos}\label{sec:videos}
We have provided some videos to augment the analysis in our paper (available in the NeurIPS supplement and at \url{http://amansinha.org/docs/OKellySiNaDuTe18_videos.zip}):

\begin{itemize}
\itemsep=5pt
\item ${\rm gail.mp4}$ provides an example of a trained GAIL model driving alongside data traces from real human drivers~\cite{Ngsim08}.
\item Example videos from rollouts. The filenames start with ``${\rm mttc=}$'' to indicate the minimum TTC that resulted between the ego and any other vehicle during the rollout. Note that even crashes have nonzero values of TTC due to the definition we used for TTC from the center of the ego vehicle (cf. Appendix \ref{section:scenario}). The videos are all played back at $2.5\times$ real-time speed. The videos included in the supplement are:
\begin{itemize}
\itemsep=5pt
\item Crashes: 
\begin{itemize}\itemsep=2pt
\item ${\rm mttc=0.23-crash.mp4}$
\item ${\rm mttc=0.30.mp4}$
\item ${\rm mttc=0.42.mp4}$
\item ${\rm mttc=0.56.mp4}$
\end{itemize}
\item Non-crashes: 
\begin{itemize}\itemsep=2pt
\item ${\rm mttc=0.23-nocrash.mp4}$
\item ${\rm mttc=0.79.mp4}$
\item ${\rm mttc=1.43.mp4}$
\item ${\rm mttc=2.01.mp4}$
\item ${\rm mttc=3.05.mp4}$
\item ${\rm mttc=6.00.mp4}$
\item ${\rm mttc=6.01.mp4}$
\item ${\rm mttc=10.11.mp4}$
\end{itemize}
\end{itemize}
These videos contain overhead, RGB, segmented, and depth views. We also include higher-resolution RGB videos with the same base names as above but the extension ``${\rm \_hires.mp4}$''.
\end{itemize} 
\end{document}